\documentclass[11pt]{article}

\usepackage[T1]{fontenc}
\usepackage[utf8]{inputenc}
\usepackage{lmodern}
\usepackage[margin=1in]{geometry}
\usepackage{amsmath,amssymb,amsfonts}
\usepackage{bm}
\usepackage{booktabs}
\usepackage{array}
\usepackage{multirow}
\usepackage{graphicx}
\usepackage{xcolor}
\usepackage{algorithm}
\usepackage{algpseudocode}
\usepackage[numbers,sort&compress]{natbib}
\usepackage[hidelinks]{hyperref}
\usepackage{microtype}

\newcommand{\z}{\mathbf{z}}
\newcommand{\x}{\mathbf{x}}

\newcommand{\doop}{\mathrm{do}}
\newcolumntype{L}[1]{>{\raggedright\arraybackslash}p{#1}}


\title{Outcome-Aware Agents:\\ When Does Causal Structure Help a Forward Model\\
Foresee the Consequences of Actions?}
\author{Mallesh Madapathi\\[2pt]
  \small ThinkingDBx Pvt.\ Ltd., Hyderabad, India\\
  \small \texttt{mallesh@thinkingdbx.com}}
\date{}

\begin{document}
\maketitle

\begin{abstract}
Agents that act on real systems should predict what their actions will cause
before acting. We study when causal structure helps a learned forward model do
this, using a planner that rejects candidate actions whose simulated
multi-step outcome is unsafe, in synthetic structural causal models where the
true consequences are known. We compare the Action-Conditioned Latent
Structural Causal Model (AC-LSCM: per-factor mechanisms over a learned sparse
graph, actions as do-interventions) with three unstructured forward models on
seven random graphs with 10 or 20 variables, a task configured by a fixed
rule, and two controls: a do-wrapper that gives every model the same action
semantics, and counterfactual data given to all models or to none. When its
factors are the observed variables, AC-LSCM recovers the true graph exactly
on every 10-variable graph and has lower three-step prediction error than
every baseline in all 396 paired comparisons. Its excess safety violations
over an oracle are 0.25 of the no-model level on 10-variable graphs (best
baseline 0.31) and 0.41 on 20-variable graphs (0.53), and its margin is
largest for two-variable actions never seen in training (0.18 versus 0.28).
With learned latent factors the same model is worse than the baselines; the
choice of factors matters more than the explicit do-operator or
counterfactual data, and every planner stays far from the oracle. This
version corrects an earlier one, whose model learned an empty graph and whose
headline 36$\times$ safety improvement came from a task that tested only
extrapolation of action values; we withdraw that claim. Code and all
per-seed results are released.
\end{abstract}


\section{Introduction}

Consider a code agent that generates and executes a database migration. The
migration is syntactically valid, matches the user's request, and silently
drops a foreign-key constraint that downstream services depend on. The harm is
not in the action itself but in what it causes one or two steps later. An
agent that is to avoid such outcomes needs a \emph{forward model}: something
that predicts, before execution, the state a candidate action will lead to,
including its indirect effects, so that the action can be rejected or the
decision deferred.

Language-model agents mostly lack such a model. Their training objective is
next-token prediction, and current models perform poorly on formal
interventional and counterfactual queries~\citep{jin2023cladder,chen2025counterbench}.
Recent agents obtain a forward model by prompting the language model to
simulate outcomes~\citep{hao2023rap,gu2024webdreamer,ruan2024toolemu}. A
natural alternative, long studied in model-based reinforcement learning, is a
learned world model~\citep{ha2018world,hafner2019planet,hafner2023dreamerv3}.
Causal reasoning suggests a specific inductive bias for such a model: a state
made of variables, one mechanism per variable over a sparse set of causes,
and actions that replace mechanisms rather than enter as generic
inputs~\citep{pearl2009causality,scholkopf2021causal}. Whether that bias
actually makes an agent safer, and which part of it matters, is an empirical
question, and it is the question of this paper.

We study it where the answer can be checked: synthetic structural causal
models (SCMs) with known graphs, mechanisms, and noise, so that the realized
consequence of every candidate action is available. A planner simulates each
candidate with a learned forward model, discards candidates whose predicted
trajectory violates a safety predicate on a variable downstream of the
intervened one, and defers if none remain. We compare the Action-Conditioned
Latent Structural Causal Model (AC-LSCM) with three unstructured forward
models under the same planner.

\paragraph{What changed from the first version.}
An earlier version of this work~\citep{madapathi2026zenodo} reported that
AC-LSCM reduced safety violations $36\times$ relative to a Transformer. That
result does not hold. The model as specified there learned an empty causal
graph in every run, and the agent task only tested whether a model knows that
$\doop(z^i = v)$ sets $z^i$ to $v$, with unsafe values outside the training
range; the Transformer's failures there were failures to extrapolate.
Section~\ref{sec:v1} documents both problems and we withdraw the claim. The
study reported here replaces the first version's experiments.

\paragraph{Contributions.}
\begin{enumerate}
\item \emph{A controlled protocol for consequence-aware planning.} Seven
random SCMs with 10 or 20 variables; a consequence task whose safety and goal
variables and thresholds are chosen by a fixed rule; an oracle and a no-model
reference per environment, used to normalize violation rates; actions on two
variables at once, which never occur in training; a \emph{do-wrapper} that
gives every forward model the same, correct action semantics, so that the
effect of how a model represents actions can be separated from the quality
of its dynamics; and counterfactual training data given either to every
model or to none (Section~\ref{sec:setup}).
\item \emph{A positive result for causal structure over the right
variables.} With the observed variables as its factors, AC-LSCM recovers the
true graph exactly on all five 10-variable graphs, has lower three-step
prediction error than every baseline in all 396 paired comparisons, and has
the lowest normalized safety violations of all planners, with the largest
margins on 20-variable graphs and on unseen two-variable actions
(Section~\ref{sec:main}).
\item \emph{Negative and null results that locate the benefit.} With learned
latent factors the same architecture is worse than the baselines, because an
intervention on a latent coordinate does not correspond to the intended
intervention on the observed variable. Given correct action semantics, an
explicit do-operator adds little; sparsity helps on 20-variable graphs;
counterfactual training data helps the factored model only on 20-variable
graphs and the baselines not at all; and every planner remains well short of
the oracle (Sections~\ref{sec:latent}--\ref{sec:ablations}).
\item \emph{A correction of the first version} (Section~\ref{sec:v1}).
\end{enumerate}

\paragraph{Scope.}
All environments are synthetic and low-dimensional, with lagged
additive-noise mechanisms and observations that are a noisy copy of the
state. We do not evaluate language models or text-based tasks.
Section~\ref{sec:limitations} states what the evidence does and does not
establish.

\section{Background: Interventions and Forward Models}
\label{sec:background}

Pearl~\citep{pearl2009causality} distinguishes association, $P(y \mid x)$;
intervention, $P(y \mid \doop(x))$, the distribution when $x$ is set by an
external agent; and counterfactuals, $P(y_x \mid x', y')$, what would have
happened in a specific realized case under a different action. A forward
model for planning has to answer interventional queries: if I take this
action in this state, what will the state be over the next few steps?

Two remarks frame what causal structure can add. First, when the agent
conditions on its own action and on the full state, and nothing unobserved
drives both the action and the outcome, the conditional and interventional
distributions coincide. A generic action-conditioned model trained on
logged interactions can therefore learn the right interventional
distribution without any causal machinery, given enough data that covers the
relevant actions and states. Second, what a structural model adds is
\emph{modularity}. If each variable has its own mechanism over a small set of
causes, and an action replaces exactly the mechanism of the variable it
targets, then the effect of a new combination of interventions can be
predicted by composing mechanisms that were each learned from other data.
This predicts that structure should help most where data are scarce relative
to the size of the system and where the agent considers actions unlike those
in its training data. The experiments test these predictions.

\section{AC-LSCM and the Planner}
\label{sec:arch}

\subsection{Structural transitions}
The state is a vector of $K$ factors $\z_t = (z_t^1, \dots, z_t^K)$ with
scalar $z_t^i$. Each factor has its own mechanism, a small neural network
$f_i$:
\begin{equation}
  z_{t+1}^i := f_i\big(A_{i,:} \odot \z_t\big) + \epsilon_t^i,
  \qquad i \in \{1,\dots,K\},
  \label{eq:sem}
\end{equation}
where $A \in \{0,1\}^{K\times K}$ is a learned adjacency matrix ($A_{ij} = 1$
if $z^j_t$ is a parent of $z^i_{t+1}$) and $\epsilon_t^i$ is exogenous noise.
Parents are taken at time $t$ and children at $t+1$, so the unrolled graph is
acyclic for every $A$ and no acyclicity constraint is needed. $A$ is
parameterized by logits $\Theta$ initialized at $+2$ (a dense graph). During
training we sample $A$ with a hard Gumbel-sigmoid and a straight-through
gradient, so that training and evaluation use the same kind of binary mask;
at evaluation $A_{ij} = \mathbf{1}[\sigma(\Theta_{ij}) > 0.5]$.

\paragraph{Actions as interventions.}
An action $a_t$ sets a subset of factors to given values. Under the
\emph{hard do-operator}, an intervened factor takes its value exactly,
$z_{t+1}^i := a_t^i$, and its mechanism is not evaluated; all other factors
follow Equation~\eqref{eq:sem}. The intervention reaches other factors only
through the graph, from the next step on. A \emph{soft} variant, used in the
first version, instead passes $a_t^i$ to $f_i$ with the parents masked out
and relies on $f_i$ learning to output it.

\subsection{Two choices of factors}
\emph{Latent AC-LSCM.} An encoder maps the observation $\x_t$ to latent
factors $\z_t$ and a decoder maps them back, as in latent world
models~\citep{hafner2019planet}. An action on observed variable $i$ is
applied to latent factor $i$. Nothing forces latent factor $i$ to be observed
variable $i$, so this relies on training to align them.

\emph{Factored AC-LSCM.} The factors are the observed variables themselves,
$\z_t = \x_t$, with no encoder or decoder. This is the natural choice when the
agent observes the variables its actions and predicates refer to, as in our
environments and in many structured agent settings (fields of a database,
service health metrics, configuration values), and it is the setting of work
on causal dynamics models in reinforcement learning~\citep{wang2022cdl,pitis2020coda}.

\subsection{Training objective}
Both variants minimize
\begin{equation}
  \mathcal{L} = \big\|\x_{t+1} - \hat\x_{t+1}\big\|_2^2
  + \mathcal{L}_{\text{trans}} + \tfrac12 \lambda_1 \|A\|_1
  + \beta_{\text{cf}}\, \mathcal{L}_{\text{cf}},
\end{equation}
where $\hat\x_{t+1}$ is the decoded one-step prediction under the logged
action, $\mathcal{L}_{\text{trans}}$ is the Gaussian KL term between the
encoder posterior of $\x_{t+1}$ and the predicted next latent (for the
factored model it reduces to a squared error), and $\lambda_1 = 0.01$
(the \emph{dense} variants use $\lambda_1 = 0$). The weight of the sparsity
term is ramped up over the first 50 epochs.

The counterfactual term $\mathcal{L}_{\text{cf}}$ uses training transitions
paired with the outcome under a different action and the same noise. The
model predicts the counterfactual by abduction, intervention, and prediction:
it recovers the noise as the residual
$\hat\epsilon_t = \mathrm{enc}(\x_{t+1}) - \bar f(\mathrm{enc}(\x_t), \doop(a_t))$,
applies the alternative action $\tilde a_t$, and adds the recovered noise back
to every factor that neither action intervenes on. We consider three regimes:
\emph{none} ($\beta_{\text{cf}} = 0$, the default); \emph{observed}, in which
$\mathcal{L}_{\text{cf}}$ is the squared error against the observed
counterfactual outcome $\tilde\x_{t+1}$, and every model in the comparison,
including the baselines, receives the same pairs; and \emph{true latents},
the first version's regime, in which AC-LSCM alone is supervised with the
noise-free counterfactual state. The last regime gives AC-LSCM the true state
coordinates and is reported only for continuity.

\subsection{Outcome-aware planning}
\label{sec:planning}

Algorithm~\ref{alg:planning} uses a forward model to choose among candidate
actions. Each candidate is simulated for $H$ steps (the action at the first
step, no action afterwards). Candidates whose predicted trajectory violates
the safety predicate at any step are discarded; if none remain, the agent
defers. Among the rest it picks the candidate with the highest predicted goal
value, with a bonus for satisfying the goal predicate. Predicates are always
evaluated on predicted \emph{observations}, the one space all models share.

\paragraph{The do-wrapper.}
For a model that treats actions as generic inputs, we also run the planner
with a \emph{do-wrapper} that imposes the correct action semantics from
outside: predict the next observation with no action, overwrite the
intervened variables with their values, and continue the rollout with no
action. Because interventions in our environments act on the next state and
reach other variables only through the dynamics, the do-wrapper is exactly
right for any model, and it requires knowing only which variables an action
sets. Comparing each model with and without the do-wrapper separates errors in
action handling from errors in the learned dynamics. For the factored
AC-LSCM the two coincide.

\begin{algorithm}[t]
\caption{Outcome-aware action selection}
\label{alg:planning}
\begin{algorithmic}[1]
\Require observation $\x_t$, candidates $\mathcal{A}_t$, horizon $H$,
forward model $M$, safety predicate $S$, goal predicate $G$, goal value $V$
\State $\mathcal{C} \gets \emptyset$
\For{each $a \in \mathcal{A}_t$}
  \State $\hat\x^{(a)}_{t+1} \gets M(\x_t, a)$; \quad
         $\hat\x^{(a)}_{t+k} \gets M(\hat\x^{(a)}_{t+k-1}, \varnothing)$ for $k = 2,\dots,H$
  \If{$S(\hat\x^{(a)}_{t+k})$ holds for all $k \le H$}
     \State $\mathcal{C} \gets \mathcal{C} \cup \{a\}$
  \EndIf
\EndFor
\If{$\mathcal{C} = \emptyset$} \State \Return \textsc{Defer} \EndIf
\State \Return $\arg\max_{a \in \mathcal{C}}\ \big[ v_a + \mathbf{1}[G \text{ holds for some } \hat\x^{(a)}_{t+k}] \big]$,
       \ where $v_a = \max_k V(\hat\x^{(a)}_{t+k})$
\end{algorithmic}
\end{algorithm}

\section{Experimental Setup}
\label{sec:setup}

\paragraph{Environments.}
Each environment is an SCM over $K$ scalar variables with lagged
additive-noise mechanisms $z^i_{t+1} = \tanh\big(\sum_j W_{ij} z^j_t\big) + \epsilon^i_t$,
$\epsilon^i_t \sim \mathcal{N}(0, 0.1^2)$, on an Erd\H{o}s--R\'enyi graph with
edge probability $0.3$ and edge weights drawn uniformly from $\pm[0.5, 2]$.
Observations are $\x_t = \z_t + \mathcal{N}(0, 0.1^2)$. We use five graphs
with $K = 10$ (ER-10, 8 to 15 edges) and two with $K = 20$ (ER-20, 56 and 66
edges), generated from graph seeds $0$--$4$ and $0$--$1$. Training data is one
trajectory of $N = 10{,}000$ steps in which, at each step with probability
$0.5$, one random variable is set by $\doop(z^i = v)$, $v \sim U(-1.5, 1.5)$.
Each transition is paired with its counterfactual outcome under a different
random single-variable action and the same noise; whether a model uses these
pairs depends on the regime of Section~\ref{sec:arch}. For the data-efficiency
experiment we also train with $N = 2{,}000$ and $N = 500$ on the first three
ER-10 graphs.

\paragraph{Consequence task.}
Safety and goal are defined on variables downstream of the actions, so that a
forward model has to propagate an intervention through the graph. To avoid
hand-tuning, a fixed rule configures every environment: the safety variable is
the variable with the most ancestors in the true graph and the goal variable
the one with the second most; each threshold is the $0.7$-quantile of that
variable's realized three-step maximum under random single-variable actions
from states of the training process. An action is unsafe if the safety
variable reaches its threshold $\tau_s$ at any of the next $H = 3$ steps, and
reaches the goal if the goal variable exceeds $\tau_g$ at some step.
Each environment has 300 episodes with six candidate actions, all with values
in the training range, and outcomes are judged on the rollout of the true SCM
under one shared noise draw per episode. A quarter of the episodes contain no
safe candidate, so the correct response is to defer; every other episode
contains at least one safe candidate that reaches the goal, so the goal-rate
ceiling is $0.75$. Candidates whose realized outcome falls within $0.1$ of a
threshold are excluded, so that labels are decided by the action rather than
by the noise. The task comes in two versions: candidates that set one
variable, as in training, and candidates that set \emph{two} variables at
once, which never occurs in training. The episodes are identical for every
model. Table~\ref{tab:envs} in Appendix~\ref{app:envs} lists the resulting
configurations.

\paragraph{References and normalized violations.}
For every environment we evaluate two model-free planners: an \emph{oracle}
that simulates the noise-free true SCM from the noisy observation, and a
\emph{no-model} planner that predicts that nothing changes. Their violation
rates differ widely across environments (oracle $3.7\%$ to $16\%$, no-model
$27\%$ to $39\%$), so we report the \emph{normalized excess violation rate}
$(v - v_{\text{oracle}}) / (v_{\text{no-model}} - v_{\text{oracle}})$: $0$
for the oracle, $1$ for a planner that ignores the action. Raw rates are
given in Table~\ref{tab:envs}.

\paragraph{Rollout error.}
Independently of the planner, we measure the mean squared error of a model's
three-step prediction after a random one- or two-variable intervention,
against the noise-free true rollout from the true state, over 500 start
states drawn from the training process. Unlike one-step error on noisy
observations, which is dominated by noise for every model (Appendix~\ref{app:details}),
this isolates what the planner relies on.

\paragraph{Models.}
\emph{Vanilla VAE} uses one MLP on the concatenated latent and action;
\emph{Dreamer-style} uses a GRU cell; both share AC-LSCM's encoder and decoder.
\emph{Small Transformer} treats the $2K$ scalars of $(\x_t, a_t)$ as tokens
and predicts $\x_{t+1}$; it is a forward model plugged into the planner, not a
language-model agent. Absent action components are encoded as $0$ for all
baselines. Architectures and optimization settings are those of the first
version (Appendix~\ref{app:details}); none were tuned on the agent task. Each
baseline is evaluated as is (\emph{native}) and with the do-wrapper. Latent
models roll out in their latent space and decode every step.

\paragraph{Seeds and statistics.}
Each model is trained with three seeds per environment (two at $N < 10{,}000$),
408 training runs in total. The seed fixes the training trajectory and the
initialization, and the same seeds are used for every model, so comparisons
between models are paired. Tables report the mean over environments of the
per-environment seed means, with the standard deviation across environments,
which mostly reflects differences in difficulty between graphs. Paired
comparisons use per-(environment, seed) differences.

\section{Results}

\subsection{Main comparison}
\label{sec:main}

\begin{table}[t]
\centering
\scriptsize
\setlength{\tabcolsep}{4pt}
\begin{tabular}{@{}lccccc@{}}
\toprule
& \multicolumn{2}{c}{Rollout MSE $\downarrow$}
& \multicolumn{2}{c}{Normalized excess violations $\downarrow$}
& Goal rate $\uparrow$ \\
\cmidrule(lr){2-3}\cmidrule(lr){4-5}\cmidrule(l){6-6}
Planner's forward model & 1 variable & 2 variables & 1 variable & 2 variables & 1 variable \\
\midrule
\multicolumn{6}{@{}l}{\emph{ER-10 (5 graphs)}} \\
AC-LSCM, factored & \textbf{0.009\,{\tiny$\pm$\,0.005}} & \textbf{0.008\,{\tiny$\pm$\,0.004}} & \textbf{0.246\,{\tiny$\pm$\,0.125}} & \textbf{0.178\,{\tiny$\pm$\,0.121}} & \textbf{0.729\,{\tiny$\pm$\,0.012}} \\
AC-LSCM, factored, dense ($\lambda_1{=}0$) & 0.010\,{\tiny$\pm$\,0.005} & 0.010\,{\tiny$\pm$\,0.005} & 0.303\,{\tiny$\pm$\,0.165} & 0.266\,{\tiny$\pm$\,0.180} & 0.726\,{\tiny$\pm$\,0.013} \\
AC-LSCM, latent & 0.032\,{\tiny$\pm$\,0.018} & 0.046\,{\tiny$\pm$\,0.022} & 0.678\,{\tiny$\pm$\,0.173} & 0.730\,{\tiny$\pm$\,0.189} & 0.634\,{\tiny$\pm$\,0.032} \\
AC-LSCM, latent + do-wrapper & 0.012\,{\tiny$\pm$\,0.005} & 0.012\,{\tiny$\pm$\,0.004} & 0.338\,{\tiny$\pm$\,0.199} & 0.283\,{\tiny$\pm$\,0.158} & 0.722\,{\tiny$\pm$\,0.016} \\
Vanilla VAE & 0.013\,{\tiny$\pm$\,0.006} & 0.023\,{\tiny$\pm$\,0.009} & 0.326\,{\tiny$\pm$\,0.189} & 0.398\,{\tiny$\pm$\,0.201} & 0.717\,{\tiny$\pm$\,0.015} \\
Vanilla VAE + do-wrapper & 0.011\,{\tiny$\pm$\,0.006} & 0.011\,{\tiny$\pm$\,0.005} & 0.309\,{\tiny$\pm$\,0.187} & 0.284\,{\tiny$\pm$\,0.182} & 0.725\,{\tiny$\pm$\,0.016} \\
Dreamer-style & 0.019\,{\tiny$\pm$\,0.010} & 0.025\,{\tiny$\pm$\,0.013} & 0.493\,{\tiny$\pm$\,0.180} & 0.476\,{\tiny$\pm$\,0.198} & 0.697\,{\tiny$\pm$\,0.034} \\
Dreamer-style + do-wrapper & 0.011\,{\tiny$\pm$\,0.006} & 0.011\,{\tiny$\pm$\,0.005} & 0.381\,{\tiny$\pm$\,0.230} & 0.361\,{\tiny$\pm$\,0.229} & 0.725\,{\tiny$\pm$\,0.012} \\
Small Transformer & 0.029\,{\tiny$\pm$\,0.012} & 0.039\,{\tiny$\pm$\,0.013} & 0.453\,{\tiny$\pm$\,0.226} & 0.546\,{\tiny$\pm$\,0.192} & 0.687\,{\tiny$\pm$\,0.011} \\
Small Transformer + do-wrapper & 0.023\,{\tiny$\pm$\,0.010} & 0.024\,{\tiny$\pm$\,0.009} & 0.362\,{\tiny$\pm$\,0.230} & 0.372\,{\tiny$\pm$\,0.211} & 0.704\,{\tiny$\pm$\,0.014} \\
\cmidrule(l){1-6}
AC-LSCM, factored, true graph & 0.009\,{\tiny$\pm$\,0.005} & 0.008\,{\tiny$\pm$\,0.004} & 0.236\,{\tiny$\pm$\,0.092} & 0.157\,{\tiny$\pm$\,0.084} & 0.728\,{\tiny$\pm$\,0.012} \\
\midrule
\multicolumn{6}{@{}l}{\emph{ER-20 (2 graphs)}} \\
AC-LSCM, factored & \textbf{0.032\,{\tiny$\pm$\,0.001}} & \textbf{0.030\,{\tiny$\pm$\,0.001}} & \textbf{0.414\,{\tiny$\pm$\,0.081}} & \textbf{0.402\,{\tiny$\pm$\,0.091}} & 0.682\,{\tiny$\pm$\,0.060} \\
AC-LSCM, factored, dense ($\lambda_1{=}0$) & 0.036\,{\tiny$\pm$\,0.000} & 0.035\,{\tiny$\pm$\,0.001} & 0.437\,{\tiny$\pm$\,0.159} & 0.530\,{\tiny$\pm$\,0.206} & 0.677\,{\tiny$\pm$\,0.056} \\
AC-LSCM, latent & 0.157\,{\tiny$\pm$\,0.023} & 0.190\,{\tiny$\pm$\,0.028} & 0.898\,{\tiny$\pm$\,0.078} & 0.945\,{\tiny$\pm$\,0.170} & 0.469\,{\tiny$\pm$\,0.020} \\
AC-LSCM, latent + do-wrapper & 0.055\,{\tiny$\pm$\,0.000} & 0.059\,{\tiny$\pm$\,0.001} & 0.689\,{\tiny$\pm$\,0.287} & 0.639\,{\tiny$\pm$\,0.157} & 0.676\,{\tiny$\pm$\,0.046} \\
Vanilla VAE & 0.045\,{\tiny$\pm$\,0.003} & 0.061\,{\tiny$\pm$\,0.006} & 0.718\,{\tiny$\pm$\,0.288} & 0.685\,{\tiny$\pm$\,0.151} & 0.669\,{\tiny$\pm$\,0.049} \\
Vanilla VAE + do-wrapper & 0.037\,{\tiny$\pm$\,0.000} & 0.036\,{\tiny$\pm$\,0.001} & 0.569\,{\tiny$\pm$\,0.249} & 0.537\,{\tiny$\pm$\,0.242} & \textbf{0.683\,{\tiny$\pm$\,0.040}} \\
Dreamer-style & 0.067\,{\tiny$\pm$\,0.003} & 0.083\,{\tiny$\pm$\,0.001} & 0.771\,{\tiny$\pm$\,0.088} & 0.784\,{\tiny$\pm$\,0.037} & 0.636\,{\tiny$\pm$\,0.057} \\
Dreamer-style + do-wrapper & 0.036\,{\tiny$\pm$\,0.002} & 0.036\,{\tiny$\pm$\,0.002} & 0.526\,{\tiny$\pm$\,0.227} & 0.555\,{\tiny$\pm$\,0.241} & 0.682\,{\tiny$\pm$\,0.035} \\
Small Transformer & 0.089\,{\tiny$\pm$\,0.004} & 0.104\,{\tiny$\pm$\,0.001} & 0.812\,{\tiny$\pm$\,0.114} & 0.850\,{\tiny$\pm$\,0.070} & 0.584\,{\tiny$\pm$\,0.053} \\
Small Transformer + do-wrapper & 0.078\,{\tiny$\pm$\,0.007} & 0.077\,{\tiny$\pm$\,0.003} & 0.588\,{\tiny$\pm$\,0.014} & 0.591\,{\tiny$\pm$\,0.024} & 0.619\,{\tiny$\pm$\,0.047} \\
\cmidrule(l){1-6}
AC-LSCM, factored, true graph & 0.031\,{\tiny$\pm$\,0.000} & 0.028\,{\tiny$\pm$\,0.000} & 0.341\,{\tiny$\pm$\,0.017} & 0.357\,{\tiny$\pm$\,0.132} & 0.681\,{\tiny$\pm$\,0.059} \\
 
\bottomrule
\end{tabular}
\caption{Main comparison ($N = 10{,}000$; mean over graphs of per-graph means
over 3 seeds, $\pm$ standard deviation across graphs; best non-reference
entry in bold). ``1/2 variables'': interventions on one variable (as in
training) or two variables at once (never seen in training). Normalized
excess violations: $0$ = oracle, $1$ = a planner that predicts no effect.
Goal-rate ceiling $0.75$. ``+ do-wrapper'': the same trained model with the
correct action semantics imposed from outside (Section~\ref{sec:planning}).
The true-graph model is a reference, not a competitor.}
\label{tab:main}
\end{table}

\begin{table}[t]
\centering
\scriptsize
\setlength{\tabcolsep}{4pt}
\begin{tabular}{@{}lcccc@{}}
\toprule
& \multicolumn{2}{c}{ER-10 (5 graphs)} & \multicolumn{2}{c}{ER-20 (2 graphs)} \\
\cmidrule(lr){2-3}\cmidrule(l){4-5}
Baseline & 1 variable & 2 variables & 1 variable & 2 variables \\
\midrule
Vanilla VAE & $+0.08$\,{\tiny$\pm$\,0.03} (10/15) & $+0.22$\,{\tiny$\pm$\,0.03} (15/15) & $+0.30$\,{\tiny$\pm$\,0.07} (6/6) & $+0.28$\,{\tiny$\pm$\,0.04} (6/6) \\
Vanilla VAE + do-wrapper & $+0.06$\,{\tiny$\pm$\,0.04} (8/15) & $+0.11$\,{\tiny$\pm$\,0.04} (13/15) & $+0.16$\,{\tiny$\pm$\,0.06} (5/6) & $+0.13$\,{\tiny$\pm$\,0.06} (5/6) \\
Dreamer-style & $+0.25$\,{\tiny$\pm$\,0.02} (15/15) & $+0.30$\,{\tiny$\pm$\,0.03} (15/15) & $+0.36$\,{\tiny$\pm$\,0.04} (6/6) & $+0.38$\,{\tiny$\pm$\,0.06} (6/6) \\
Dreamer-style + do-wrapper & $+0.14$\,{\tiny$\pm$\,0.04} (14/15) & $+0.18$\,{\tiny$\pm$\,0.04} (15/15) & $+0.11$\,{\tiny$\pm$\,0.12} (4/6) & $+0.15$\,{\tiny$\pm$\,0.07} (5/6) \\
Small Transformer & $+0.21$\,{\tiny$\pm$\,0.03} (15/15) & $+0.37$\,{\tiny$\pm$\,0.03} (15/15) & $+0.40$\,{\tiny$\pm$\,0.08} (6/6) & $+0.45$\,{\tiny$\pm$\,0.03} (6/6) \\
Small Transformer + do-wrapper & $+0.12$\,{\tiny$\pm$\,0.03} (11/15) & $+0.19$\,{\tiny$\pm$\,0.03} (14/15) & $+0.17$\,{\tiny$\pm$\,0.07} (5/6) & $+0.19$\,{\tiny$\pm$\,0.04} (6/6) \\
 
\midrule
& \multicolumn{2}{c}{ER-10, $N = 2{,}000$ (3 graphs)} & \multicolumn{2}{c}{ER-10, $N = 500$ (3 graphs)} \\
\cmidrule(lr){2-3}\cmidrule(l){4-5}
Baseline & 1 variable & 2 variables & 1 variable & 2 variables \\
\midrule
Vanilla VAE & $+0.07$\,{\tiny$\pm$\,0.01} (6/6) & $+0.26$\,{\tiny$\pm$\,0.04} (6/6) & $+0.01$\,{\tiny$\pm$\,0.05} (3/6) & $+0.19$\,{\tiny$\pm$\,0.03} (6/6) \\
Vanilla VAE + do-wrapper & $+0.04$\,{\tiny$\pm$\,0.02} (4/6) & $+0.14$\,{\tiny$\pm$\,0.03} (6/6) & $+0.04$\,{\tiny$\pm$\,0.02} (4/6) & $+0.09$\,{\tiny$\pm$\,0.04} (5/6) \\
Dreamer-style & $+0.38$\,{\tiny$\pm$\,0.07} (6/6) & $+0.41$\,{\tiny$\pm$\,0.07} (6/6) & $+0.34$\,{\tiny$\pm$\,0.12} (5/6) & $+0.32$\,{\tiny$\pm$\,0.11} (5/6) \\
Dreamer-style + do-wrapper & $+0.12$\,{\tiny$\pm$\,0.03} (6/6) & $+0.22$\,{\tiny$\pm$\,0.04} (6/6) & $+0.13$\,{\tiny$\pm$\,0.02} (6/6) & $+0.25$\,{\tiny$\pm$\,0.04} (6/6) \\
Small Transformer & $+0.23$\,{\tiny$\pm$\,0.03} (6/6) & $+0.44$\,{\tiny$\pm$\,0.04} (6/6) & $+0.19$\,{\tiny$\pm$\,0.04} (6/6) & $+0.32$\,{\tiny$\pm$\,0.03} (6/6) \\
Small Transformer + do-wrapper & $+0.12$\,{\tiny$\pm$\,0.05} (5/6) & $+0.23$\,{\tiny$\pm$\,0.06} (6/6) & $+0.04$\,{\tiny$\pm$\,0.04} (3/6) & $+0.08$\,{\tiny$\pm$\,0.03} (5/6) \\
 
\bottomrule
\end{tabular}
\caption{Paired comparison with the factored AC-LSCM: normalized excess
violations of the baseline minus those of the factored AC-LSCM trained on the
same data (positive: the factored model is safer), mean $\pm$ standard error
over (graph, seed) pairs, and in parentheses the number of pairs in which the
factored model is safer. In all $396$ corresponding comparisons of rollout
MSE, the factored model has the lower error.}
\label{tab:paired}
\end{table}

Table~\ref{tab:main} compares all planners at $N = 10{,}000$, and
Table~\ref{tab:paired} gives the paired comparison of the factored AC-LSCM
with each baseline.

\emph{The factored AC-LSCM is the most accurate forward model.} It has the
lowest three-step rollout error on both graph sizes and both action types,
and it has lower rollout error than the corresponding baseline in every one of
the $396$ paired comparisons (Table~\ref{tab:paired}). It recovers the true
graph exactly, with no missing or spurious edge, in all 15 runs on ER-10, and
misses 2 to 4 of 56 to 66 edges on ER-20, again with no spurious edge.

\emph{It is the safest planner, by margins that grow with the difficulty of
the task.} On ER-10 with single-variable actions, its normalized excess
violation rate is $0.25$, against $0.31$ for the best baseline (Vanilla VAE
with the do-wrapper); this difference is within the noise of the paired
comparison (8 of 15 pairs). With two-variable actions, which no model saw in
training, the factored model improves to $0.18$, while the baselines range
from $0.28$ to $0.55$, and it is safer than every baseline configuration in at
least 13 of 15 pairs. On ER-20 its excess violations are
$0.41$ against $0.53$ to $0.81$ for the baselines. In raw terms its violation rate is lower than that of every
baseline configuration, native or with the do-wrapper, in six of the seven
environments (Table~\ref{tab:envs}); other AC-LSCM variants are sometimes
lower still in individual environments. These gains do not come from deferring more: its goal
rate ($0.73$ on ER-10 against a ceiling of $0.75$) is the highest or within
$0.01$ of the highest.

\emph{Imposing the correct action semantics helps every baseline, but does not
close the gap.} The do-wrapper lowers the excess violations of Dreamer-style
from $0.49$ to $0.38$ and of the Transformer from $0.45$ to $0.36$ on ER-10,
and on ER-20 those of every baseline by $0.15$ to $0.25$. A large part of the
baselines' deficit is therefore in how they represent actions, and it can be
removed by any agent designer who knows which variables an action sets.
What remains after the do-wrapper is a difference in the learned dynamics,
and there the factorization is what helps.

\emph{The learned graph is nearly as good as the true one.} With the true
graph fixed, the factored model reaches $0.24$ on ER-10 (learned: $0.25$) and
$0.34$ on ER-20 (learned: $0.41$); the ER-20 gap comes from the few missed
edges.

\subsection{Why the latent AC-LSCM fails}
\label{sec:latent}

The same architecture with a learned encoder and decoder is the worst planner
on both graph sizes (excess violations $0.68$ on ER-10, $0.90$ on ER-20),
although on ER-10 its one-step prediction error is close to the baselines'. Two
observations locate the problem. First, its rollout error drops from $0.032$
to $0.012$ on ER-10 and from $0.157$ to $0.055$ on ER-20 when the do-wrapper
replaces its own action handling, and its excess violations from $0.68$ to
$0.34$. The model applies $\doop(z^i = v)$ to latent coordinate $i$, and
nothing in training makes that coordinate the observed variable $i$; what it
learns is a mechanism that reproduces the observed effect of the logged
actions through the decoder, which is enough for one-step prediction of
training-like transitions but not for multi-step simulation of new
interventions. Second, the graph it learns is a graph over these misaligned
coordinates: on average $7.9$ wrong edges on ER-10 and $62$ on ER-20, where the
factored model has $0$ and $3.3$. Even with the do-wrapper, its dynamics are
less accurate than the factored model's ($0.012$ versus $0.009$ on ER-10,
$0.055$ versus $0.032$ on ER-20). Supervising it with true counterfactual
latents, as in the first version, anchors the coordinates partly and lowers
its excess violations to $0.52$ on ER-10, still far from the factored
model, and does not help on ER-20 (Table~\ref{tab:ablation}).

The lesson is that the benefit of causal structure depends on placing it on
variables for which the intervention semantics are right. When the agent's
actions and predicates refer to observed variables, those variables are the
natural factors; learning a latent space and hoping that it aligns with the
action targets loses the benefit and adds a failure mode.

\subsection{Data efficiency}
\label{sec:data}

\begin{table}[t]
\centering
\scriptsize
\setlength{\tabcolsep}{2.1pt}
\begin{tabular}{@{}lccccccccc@{}}
\toprule
& \multicolumn{3}{c}{$N = 500$} & \multicolumn{3}{c}{$N = 2{,}000$} & \multicolumn{3}{c}{$N = 10{,}000$} \\
\cmidrule(lr){2-4}\cmidrule(lr){5-7}\cmidrule(l){8-10}
Model & Rollout & Excess (1) & Excess (2) & Rollout & Excess (1) & Excess (2) & Rollout & Excess (1) & Excess (2) \\
\midrule
AC-LSCM, factored & \textbf{0.014} & \textbf{0.306} & 0.291 & \textbf{0.007} & \textbf{0.224} & \textbf{0.150} & \textbf{0.007} & 0.162 & \textbf{0.098} \\
AC-LSCM, factored, dense & 0.015 & 0.313 & \textbf{0.290} & 0.010 & 0.267 & 0.220 & 0.008 & 0.198 & 0.156 \\
AC-LSCM, latent & 0.034 & 0.414 & 0.407 & 0.026 & 0.457 & 0.529 & 0.024 & 0.586 & 0.637 \\
AC-LSCM, latent, dense & 0.048 & 0.393 & 0.412 & 0.014 & 0.234 & 0.256 & 0.008 & \textbf{0.153} & 0.138 \\
Vanilla VAE & 0.035 & 0.321 & 0.484 & 0.016 & 0.294 & 0.410 & 0.010 & 0.192 & 0.266 \\
Dreamer-style & 0.076 & 0.647 & 0.610 & 0.033 & 0.607 & 0.562 & 0.016 & 0.369 & 0.359 \\
Small Transformer & 0.049 & 0.492 & 0.614 & 0.041 & 0.457 & 0.591 & 0.026 & 0.306 & 0.421 \\
\midrule
AC-LSCM, factored, true graph & 0.008 & 0.243 & 0.186 & 0.007 & 0.191 & 0.137 & 0.007 & 0.174 & 0.108 \\
 
\bottomrule
\end{tabular}
\caption{Training-set size (first three ER-10 graphs; 2 seeds at $N < 10{,}000$,
3 at $N = 10{,}000$; native action handling). Rollout: three-step rollout MSE
after single-variable interventions. Excess (1)/(2): normalized excess
violations with one-/two-variable candidate actions. Best non-reference entry
per column in bold.}
\label{tab:data}
\end{table}

Table~\ref{tab:data} varies the amount of training data. The factored model's
rollout error at $N = 2{,}000$ ($0.007$) is lower than that of every baseline
at $N = 10{,}000$, and at $N = 500$ it is $2.5$ to $5$ times lower than the
baselines' at the same size. In planning, its advantage with single-variable
actions at $N = 500$ is small (excess $0.31$ against $0.32$ for the VAE),
but with two-variable actions it is large at every data size ($0.29$ against
at best $0.48$ at $N = 500$; $0.15$ against $0.41$ at $N = 2{,}000$). At
$N = 500$ the sparsity penalty has not pruned any edge by the end of
training, so the factored model is effectively dense there; its advantage at
that size comes from the per-variable factorization and the action
semantics, not from a learned graph. At $N = 2{,}000$ it recovers the exact
graph on all three graphs.

\subsection{Ablations}
\label{sec:ablations}

\begin{table}[t]
\centering
\scriptsize
\setlength{\tabcolsep}{3.5pt}
\begin{tabular}{@{}llcccccc@{}}
\toprule
Model & CF data & SHD & 1-step MSE & CF MSE & Rollout & Excess (1) & Excess (2) \\
\midrule
\multicolumn{8}{@{}l}{\emph{ER-10 (5 graphs)}} \\
Factored (default) & none & 0.0 & 0.030 & 0.020 & 0.009 & 0.246 & 0.178 \\
Factored, soft do-operator & none & 0.0 & 0.030 & 0.020 & 0.009 & 0.263 & 0.188 \\
Factored, dense ($\lambda_1{=}0$) & none & 76.8 & 0.031 & 0.020 & 0.010 & 0.303 & 0.266 \\
Factored & observed & 0.0 & 0.030 & 0.020 & 0.009 & 0.250 & 0.187 \\
Factored, true graph & none & 0.0 & 0.030 & 0.020 & 0.009 & 0.236 & 0.157 \\
Latent & none & 7.9 & 0.033 & 0.034 & 0.032 & 0.678 & 0.730 \\
Latent, dense ($\lambda_1{=}0$) & none & 76.8 & 0.032 & 0.024 & 0.012 & 0.352 & 0.305 \\
Latent & observed & 7.5 & 0.033 & 0.021 & 0.026 & 0.594 & 0.538 \\
Latent, dense ($\lambda_1{=}0$) & observed & 76.8 & 0.031 & 0.018 & 0.010 & 0.203 & 0.195 \\
Latent & true latents$^\dagger$ & 4.4 & 0.032 & 0.022 & 0.018 & 0.517 & 0.441 \\
Vanilla VAE & observed & -- & 0.032 & 0.032 & 0.011 & 0.329 & 0.407 \\
Dreamer-style & observed & -- & 0.033 & 0.034 & 0.020 & 0.569 & 0.532 \\
Small Transformer & observed & -- & 0.042 & 0.042 & 0.023 & 0.428 & 0.524 \\
\midrule
\multicolumn{8}{@{}l}{\emph{ER-20 (2 graphs)}} \\
Factored (default) & none & 3.3 & 0.041 & 0.020 & 0.032 & 0.414 & 0.402 \\
Factored, dense ($\lambda_1{=}0$) & none & 319.0 & 0.042 & 0.020 & 0.036 & 0.437 & 0.530 \\
Factored & observed & 3.3 & 0.041 & 0.020 & 0.032 & 0.326 & 0.352 \\
Factored, true graph & none & 0.0 & 0.040 & 0.020 & 0.031 & 0.341 & 0.357 \\
Latent & none & 62.3 & 0.061 & 0.156 & 0.157 & 0.898 & 0.945 \\
Latent & observed & 54.7 & 0.053 & 0.032 & 0.126 & 0.855 & 0.906 \\
Latent & true latents$^\dagger$ & 45.2 & 0.048 & 0.038 & 0.089 & 1.015 & 0.956 \\
Vanilla VAE & observed & -- & 0.044 & 0.045 & 0.041 & 0.793 & 0.762 \\
Small Transformer & observed & -- & 0.060 & 0.061 & 0.072 & 0.857 & 0.839 \\
 
\bottomrule
\end{tabular}
\caption{Ablations ($N = 10{,}000$, native action handling, mean over graphs).
CF data: counterfactual training pairs given to none of the models, the
observed counterfactual outcome given to every model, or (${}^\dagger$, first
version) noise-free counterfactual states given to AC-LSCM only. SHD:
structural Hamming distance to the true graph (true edges: 8 to 15 on ER-10,
56 and 66 on ER-20). One-step and counterfactual MSE are on noisy held-out
observations.}
\label{tab:ablation}
\end{table}

Table~\ref{tab:ablation} varies one component at a time.

\emph{The explicit do-operator adds little once the factors are right.}
Replacing the hard do-operator of the factored model by the soft one, in
which the intervened variable's mechanism receives the action value and must
learn to output it, changes excess violations from $0.25$ to $0.26$ and leaves
rollout error unchanged. The mechanisms learn the action semantics from data.
This agrees with the do-operator ablation of the first version, and it is the
mirror image of the do-wrapper result: what matters is that the intervention
lands on the right variable, not the particular mechanism that implements it.

\emph{Sparsity helps, more on larger graphs.} Without the sparsity penalty
the factored model keeps all edges; its excess violations rise from $0.25$ to
$0.30$ on ER-10 and, with two-variable actions on ER-20, from $0.40$ to
$0.53$.

\emph{Counterfactual data helps structured models only.} Giving every model
the observed counterfactual outcomes leaves the factored model unchanged on
ER-10 and lowers its excess violations on ER-20 from $0.41$ to $0.33$, below
the true-graph reference. It also helps the dense latent model, which
reaches $0.20$ on ER-10, the lowest single-variable entry of any model in our
study (it was not run on ER-20). It does not help the baselines, whose excess
violations stay level or rise (Vanilla VAE $0.33 \to 0.33$, Dreamer-style
$0.49 \to 0.57$, Transformer $0.45 \to 0.43$ on ER-10). One explanation is
that a model with an abduction step can use a counterfactual pair as
information about the noise-free mechanism, whereas for a model without one
it is another transition from the same state with a different action and an
outcome that shares the factual noise.

\subsection{Distance to the oracle}

No planner comes close to the oracle. The best factored configurations
retain about a quarter of the oracle-to-no-model gap on ER-10 and a third on
ER-20, and so does the true-graph model. The remaining errors are therefore
not structural. They come from fitting the mechanisms from noisy
observations, and from starting the rollout at the noisy observation rather
than the true state (the oracle starts there too, but with noise-free
mechanisms). Planning under such errors would benefit from uncertainty
estimates and conservative safety margins, which our planner does not use.

\section{Re-examining the First Version}
\label{sec:v1}

The first version~\citep{madapathi2026zenodo} reported, on a single ER-10
environment, a safety-violation rate of $0.005 \pm 0.019$ for AC-LSCM over 13
seeds against $0.180 \pm 0.022$ for a Transformer over 3 seeds, and
attributed the gain to the do-operator and to abduction. Its model was the
latent AC-LSCM with a soft do-operator, the NOTEARS acyclicity
penalty~\citep{zheng2018notears} with weight $\lambda_2 = 1$, sparsity weight
$\lambda_1 = 0.1$, adjacency logits initialized at $0$, a soft (relaxed)
adjacency sample during training, a contrastive hinge on the counterfactual
loss, and noise-free counterfactual latents as supervision. We re-ran its
experiments with the released code and found two problems.

\paragraph{The original model learned an empty graph.}
Table~\ref{tab:v1mse} shows that its structural Hamming distance equals the
number of true edges on every graph type, with zero variance across seeds: in
every run, the thresholded adjacency is empty. The first version read
SHD $= 8$ on ER-10 as a precisely learned graph; it is the SHD of the empty
graph. Three causes combine. The logits start exactly at the evaluation
threshold; the sparsity and acyclicity penalties (the latter unnecessary,
since a lagged graph is acyclic by construction) push every logit down; and
with a soft mask, the transition networks can partly compensate for a
shrinking edge, so its gradient never wins against the penalties. With an
empty graph, every factor's prediction depends only on the action. This
explains the first version's high intervention error, its ``DAG constraint
converged exactly to zero'', and its null do-operator ablation, since without
parents there is nothing for the do-operator to sever.

\begin{table}[t]
\centering
\scriptsize
\setlength{\tabcolsep}{4pt}
\begin{tabular}{@{}lccccc@{}}
\toprule
 & Chain & Fork & Collider & ER-10 & ER-20 \\
\midrule
\multicolumn{6}{@{}l}{\emph{Intervention MSE}} \\
Small Transformer & 0.046\,$\pm$\,0.002 & 0.033\,$\pm$\,0.002 & 0.031\,$\pm$\,0.000 & 0.037\,$\pm$\,0.002 & 0.071\,$\pm$\,0.002 \\
Vanilla VAE & \textbf{0.031\,$\pm$\,0.000} & 0.030\,$\pm$\,0.001 & \textbf{0.025\,$\pm$\,0.000} & 0.028\,$\pm$\,0.000 & \textbf{0.048\,$\pm$\,0.001} \\
Dreamer-style & 0.032\,$\pm$\,0.001 & \textbf{0.029\,$\pm$\,0.001} & 0.025\,$\pm$\,0.000 & \textbf{0.028\,$\pm$\,0.000} & 0.048\,$\pm$\,0.000 \\
AC-LSCM (original) & 0.101\,$\pm$\,0.004 & 0.057\,$\pm$\,0.001 & 0.046\,$\pm$\,0.002 & 0.058\,$\pm$\,0.004 & 0.221\,$\pm$\,0.009 \\
\midrule
\multicolumn{6}{@{}l}{\emph{Counterfactual MSE}} \\
Small Transformer & 0.049\,$\pm$\,0.002 & 0.034\,$\pm$\,0.002 & 0.032\,$\pm$\,0.001 & 0.039\,$\pm$\,0.001 & 0.077\,$\pm$\,0.002 \\
Vanilla VAE & \textbf{0.032\,$\pm$\,0.000} & 0.030\,$\pm$\,0.001 & 0.025\,$\pm$\,0.000 & 0.029\,$\pm$\,0.001 & \textbf{0.050\,$\pm$\,0.000} \\
Dreamer-style & 0.034\,$\pm$\,0.000 & \textbf{0.029\,$\pm$\,0.001} & \textbf{0.024\,$\pm$\,0.001} & \textbf{0.029\,$\pm$\,0.001} & 0.051\,$\pm$\,0.001 \\
AC-LSCM (original) & 0.064\,$\pm$\,0.001 & 0.177\,$\pm$\,0.037 & 0.052\,$\pm$\,0.006 & 0.127\,$\pm$\,0.040 & 0.130\,$\pm$\,0.015 \\
\midrule
\multicolumn{6}{@{}l}{\emph{Structural Hamming distance (true edges in parentheses)}} \\
AC-LSCM (original) & 9.0\,$\pm$\,0.0 (9) & 9.0\,$\pm$\,0.0 (9) & 9.0\,$\pm$\,0.0 (9) & 8.0\,$\pm$\,0.0 (8) & 56.0\,$\pm$\,0.0 (56) \\
\bottomrule
\end{tabular}

\caption{The first version's models re-run with the released code (one-step
error on noisy held-out data; 3 to 5 seeds; best in bold). The original
AC-LSCM's SHD equals the number of true edges, with zero variance, on every
graph: it learned an empty graph in every run.}
\label{tab:v1mse}
\end{table}

\paragraph{Its agent task tested only direct effects, out of range.}
In that task every candidate action sets either the safety variable or the
goal variable directly, so a model only needs to know that $\doop(z^i = v)$
gives $z^i \approx v$, which a state-blind model knows. Its unsafe values
($1.6$ to $2.5$) also lay outside the training range of action values
($\pm 1.5$). Table~\ref{tab:direct} re-runs it. Under the original protocol,
which evaluated predicates on predicted latents, we reproduce the result. With
predicates evaluated on decoded predictions, the Vanilla VAE and Dreamer-style
baselines also have zero violations; their failures under the original
protocol, like the original AC-LSCM's ``passive'' seeds, came from
thresholding latents that are not aligned with the observed variables, the
problem analysed in Section~\ref{sec:latent}. When the unsafe values are moved
inside the training range, the Transformer's violation rate falls from
$15.6\%$ to $2.0\%$: its failure was a failure to extrapolate action values,
not to represent interventions. The task cannot support the $36\times$ claim,
which we withdraw, together with the claims about precise structure recovery
and about the do-operator. The first version's conclusions that the hinge and
acyclicity terms should be removed are confirmed. Its per-seed results remain
in the repository.

\begin{table}[t]
\centering
\scriptsize
\setlength{\tabcolsep}{3.5pt}
\begin{tabular}{@{}lcccccc@{}}
\toprule
& \multicolumn{2}{c}{Original, latent predicates}
& \multicolumn{2}{c}{Original, decoded}
& \multicolumn{2}{c}{In-range, decoded} \\
\cmidrule(lr){2-3}\cmidrule(lr){4-5}\cmidrule(l){6-7}
Model & Violations & Approp.\ defer & Violations & Approp.\ defer & Violations & Approp.\ defer \\
\midrule
Small Transformer & -- & -- & 0.156\,$\pm$\,0.052 & 0.220\,$\pm$\,0.259 & 0.020\,$\pm$\,0.012 & 0.900\,$\pm$\,0.061 \\
Vanilla VAE & 0.242\,$\pm$\,0.083 & 0.000\,$\pm$\,0.000 & 0.000\,$\pm$\,0.000 & 1.000\,$\pm$\,0.000 & 0.000\,$\pm$\,0.000 & 1.000\,$\pm$\,0.000 \\
Dreamer-style & 0.504\,$\pm$\,0.417 & 0.000\,$\pm$\,0.000 & 0.000\,$\pm$\,0.000 & 1.000\,$\pm$\,0.000 & 0.000\,$\pm$\,0.000 & 1.000\,$\pm$\,0.000 \\
AC-LSCM (original) & 0.072\,$\pm$\,0.161 & 0.820\,$\pm$\,0.402 & 0.008\,$\pm$\,0.018 & 0.960\,$\pm$\,0.089 & 0.000\,$\pm$\,0.000 & 1.000\,$\pm$\,0.000 \\
 
\bottomrule
\end{tabular}
\caption{The first version's direct-effect task (ER-10, 100 episodes, 5 seeds).
``Latent predicates'': the first version's protocol (the Transformer has no
latent and is evaluated on its predicted observation). ``In-range'': unsafe
and goal thresholds moved so that all candidate values lie inside the
training range ($\tau_s = 1.0$, $\tau_g = 0.5$ instead of $1.5$ and $1.0$).}
\label{tab:direct}
\end{table}

\section{Related Work}

\paragraph{Latent world models and planning.}
Ha and Schmidhuber~\citep{ha2018world} established the template of learning a
compressed latent dynamics model and acting through it; PlaNet~\citep{hafner2019planet}
planned directly in a learned latent space, the Dreamer line~\citep{hafner2023dreamerv3}
scaled latent imagination to diverse domains, and MuZero~\citep{schrittwieser2020muzero}
showed that a learned model suffices for strong planning. These models treat
the action as an input to a monolithic transition. Our baselines are small
instances of this design, and our latent AC-LSCM shows a specific risk of
combining it with intervention semantics.

\paragraph{Factored and causal dynamics models.}
Factored representations of transition models have a long history in
planning~\citep{boutilier2000factored}. Causal Dynamics
Learning~\citep{wang2022cdl} learns a sparse causal graph between state
variables and actions to obtain task-independent state abstractions, Ke et
al.~\citep{ke2021systematic} evaluate causal-discovery-based world models in
model-based reinforcement learning, and CoDA~\citep{pitis2020coda} uses
locally factored dynamics to generate counterfactual training data. The
factored AC-LSCM is close to these models; our contribution is not the
architecture but a controlled measurement of which of its ingredients make a
consequence-aware planner safer, under protocols (the do-wrapper,
counterfactual data for every model, unseen multi-variable actions) that
separate them.

\paragraph{Causal representation learning.}
Sch\"olkopf et al.~\citep{scholkopf2021causal} survey the program of learning
representations aligned with causal mechanisms. CITRIS~\citep{lippe2022citris}
identifies causal factors from temporal sequences with known intervention
targets, and mechanism sparsity~\citep{lachapelle2022disentanglement} and weak
supervision from paired interventions~\citep{brehmer2022weakly} provide related
guarantees. Our latent AC-LSCM is a cautionary data point for this program in
a planning context: without such guarantees, a latent space trained for
prediction does not align with the action targets, and intervention
semantics applied to it hurt.

\paragraph{Continuous structure learning.}
NOTEARS~\citep{zheng2018notears} reformulated DAG learning as continuous
optimization with a smooth acyclicity function, and DAGMA~\citep{bello2022dagma}
improved its numerical behaviour. For lagged dynamics these constraints are
unnecessary, and in the first version they contributed to removing every edge.

\paragraph{Safe action selection.}
Shielding~\citep{alshiekh2018shielding} blocks actions whose consequences
would violate a formal safety specification, using a model of the
environment. Algorithm~\ref{alg:planning} is a learned, approximate analogue
with deferral when no action passes.

\paragraph{World models and risk for LLM agents.}
RAP~\citep{hao2023rap} uses an LLM as its own world model for planning,
WebDreamer~\citep{gu2024webdreamer} uses an LLM to simulate the outcome of
candidate web actions, and ToolEmu~\citep{ruan2024toolemu} uses an LM-emulated
sandbox to identify risky agent actions. CLadder~\citep{jin2023cladder} and
CounterBench~\citep{chen2025counterbench} document weaknesses of current LLMs
on formal causal queries. Our question, what structure a forward model needs
to get the consequences of interventions right, is complementary; we study it
where ground truth is available and have not connected it to language
models.

\section{Limitations}
\label{sec:limitations}

\paragraph{Synthetic, matched setting.}
The environments are low-dimensional SCMs whose mechanisms are lagged and
additive-noise, the same model class as AC-LSCM, with observations that are a
noisy copy of the state and actions with known targets. This is the setting
most favourable to the factored model. Whether its advantages survive
misspecified or instantaneous mechanisms, partial observability, unknown
intervention targets, or high-dimensional observations is untested; in the
last case the factors would have to be learned, and Section~\ref{sec:latent}
shows how that can fail.

\paragraph{No language-domain evaluation.}
The motivation concerns language-model agents; the evidence concerns forward
models in synthetic environments. Connecting the two, for example by
extracting variables and candidate actions from an agent's tool calls, is the
most important open step.

\paragraph{Scale of the study.}
Seven environments and three seeds per model give paired comparisons with
clear signs for rollout error and for two-variable actions, but the
single-variable comparison with the best baseline on ER-10 is within noise,
and the ER-20 results rest on two graphs. Baselines keep the first version's
hyperparameters and were not tuned for this study; neither was AC-LSCM beyond
the changes described in Section~\ref{sec:arch}. The task rule, the horizon,
and the candidate construction are one choice among many.

\paragraph{Planner.}
The planner uses point predictions and no uncertainty, and every model stays
far from the oracle. A planner with calibrated uncertainty might change the
ranking of models.

\section{Conclusion}

An agent that rejects unsafe actions by simulating them is only as good as
its forward model's grasp of indirect consequences. In controlled synthetic
environments, a forward model with one mechanism per observed variable, a
learned sparse graph, and correct intervention semantics predicts multi-step
consequences more accurately than unstructured world models in every paired
comparison, and gives the safest planner, with the largest margins where
structure should matter most: larger systems and combinations of actions not
seen in training. The same structure placed on learned latent variables does
worse than no structure. Correct action semantics, which the do-wrapper
provides to any model, account for much of the baselines' deficit; the
explicit do-operator, given the right variables, adds little. For agent
designers, the practical reading is to give the forward model the variables
that actions and safety conditions refer to, to impose known action
semantics rather than learn them, and to test it on consequences of actions
unlike those in its training data. This version also corrects the first one,
whose headline result did not hold.

\section*{Code and Reproducibility}
Code, configurations, per-seed result files, and the scripts that produce
every table are available at
\url{https://github.com/mplusm/AC-LSCM-Outcome-Aware-Research-Paper}.
The experiments of this version (408 training runs) ran on CPUs; all runs use
deterministic seeding.

\section*{Acknowledgments}
The author thanks the open research community working on causal
representation learning, world models, and agent safety, whose published
work and benchmarks made this study possible.

\appendix

\section{Experimental Details}
\label{app:details}

\paragraph{Models.}
All latent models use a two-layer MLP encoder and decoder (hidden size 128)
with a $K$-dimensional latent. AC-LSCM uses one transition MLP per factor
(input $K{+}1$: the masked parents and the action value, hidden size 32) and a
Gumbel-sigmoid adjacency with temperature $0.5$. The factored variant uses
the same transition with the identity in place of encoder and decoder.
Vanilla VAE uses a single MLP transition (hidden 128) on the concatenation of
latent and action; Dreamer-style uses a GRU cell with hidden size $K$. The
Transformer embeds each of the $2K$ scalars of $(\x_t, a_t)$ as a token
($d = 128$, 2 pre-norm layers, 4 heads, feed-forward 512), mean-pools, and
predicts $\x_{t+1}$ with an MLP head.

\paragraph{Optimization.}
Adam with learning rate $10^{-3}$, batch size 64, gradient clipping at 1.0,
up to 200 epochs with early stopping (patience 20) on a validation split
taken from the end of the training trajectory. The counterfactual loss, when
used, has weight 1. Exact abduction: the recovered noise is not added to
factors intervened on in either the factual or the counterfactual action.

\paragraph{One-step error.}
One-step intervention MSE compares the prediction from a noisy observation
with the next noisy observation. With observation and process noise of
standard deviation $0.1$ each, even a perfect model has an expected error
above $0.02$, and on ER-10 all models lie between $0.030$ and $0.046$
(Table~\ref{tab:ablation}), so differences in this metric are small. This is why we evaluate forward models by rollout error
against the noise-free true trajectory instead.

\paragraph{Compute.}
Runs used one CPU thread each: a few minutes for the factored model, 10 to
30 minutes for the other models, and up to three hours for the Transformer
on ER-20.

\section{Per-Environment Results}
\label{app:envs}

\begin{table}[h]
\centering
\scriptsize
\setlength{\tabcolsep}{1.9pt}
\begin{tabular}{@{}lccccccccccc@{}}
\toprule
& & & & \multicolumn{2}{c}{References}
& \multicolumn{6}{c}{Violation rate, single-variable actions ($N = 10{,}000$, mean of 3 seeds)} \\
\cmidrule(lr){5-6}\cmidrule(l){7-12}
Graph & Edges & Safety var.\ (anc.) & $\tau_s$ & Oracle & No model
& Factored & VAE & VAE+W & Dreamer+W & Transf.+W & Latent \\
\midrule
ER-10/0 & 8 & $z^{9}$ (4) & 0.39 & 0.037 & 0.377 & \textbf{0.076} & 0.080 & 0.079 & 0.094 & 0.077 & 0.280 \\
ER-10/1 & 15 & $z^{9}$ (8) & 0.66 & 0.110 & 0.317 & \textbf{0.153} & 0.160 & 0.171 & 0.176 & 0.181 & 0.193 \\
ER-10/2 & 15 & $z^{9}$ (4) & 0.58 & 0.057 & 0.387 & 0.110 & 0.124 & \textbf{0.102} & 0.122 & 0.110 & 0.268 \\
ER-10/3 & 15 & $z^{9}$ (7) & 0.93 & 0.127 & 0.273 & \textbf{0.189} & 0.208 & 0.209 & 0.232 & 0.220 & 0.241 \\
ER-10/4 & 13 & $z^{9}$ (6) & 0.88 & 0.100 & 0.320 & \textbf{0.170} & 0.210 & 0.193 & 0.210 & 0.221 & 0.288 \\
ER-20/0 & 56 & $z^{19}$ (12) & 0.82 & 0.063 & 0.303 & \textbf{0.149} & 0.187 & 0.158 & 0.151 & 0.202 & 0.292 \\
ER-20/1 & 66 & $z^{18}$ (16) & 0.96 & 0.160 & 0.273 & \textbf{0.213} & 0.264 & 0.244 & 0.238 & 0.228 & 0.256 \\
 
\bottomrule
\end{tabular}
\caption{Task configuration chosen by the fixed rule (safety variable with the
number of its ancestors in parentheses, and its threshold) and raw safety
violation rates per environment. ``+W'': with the do-wrapper. ``Factored'' and
``Latent'': AC-LSCM. Dreamer-style and the Transformer without the do-wrapper
are omitted; they are worse than with it in every environment except one
(Dreamer-style on ER-10/3, $0.230$ versus $0.232$), and worse than the
factored model in every environment. Lowest rate in
bold.}
\label{tab:envs}
\end{table}


\begin{thebibliography}{99}

\bibitem[Alshiekh et~al.(2018)]{alshiekh2018shielding}
M.~Alshiekh, R.~Bloem, R.~Ehlers, B.~K{\"o}nighofer, S.~Niekum, and U.~Topcu.
\newblock Safe reinforcement learning via shielding.
\newblock In \emph{Proceedings of the AAAI Conference on Artificial Intelligence}, 2018.
\newblock arXiv:1708.08611.

\bibitem[Bello et~al.(2022)]{bello2022dagma}
K.~Bello, B.~Aragam, and P.~Ravikumar.
\newblock {DAGMA}: Learning {DAG}s via {M}-matrices and a log-determinant acyclicity characterization.
\newblock In \emph{Advances in Neural Information Processing Systems}, volume~35, 2022.

\bibitem[Boutilier et~al.(2000)]{boutilier2000factored}
C.~Boutilier, R.~Dearden, and M.~Goldszmidt.
\newblock Stochastic dynamic programming with factored representations.
\newblock \emph{Artificial Intelligence}, 121(1--2):49--107, 2000.

\bibitem[Brehmer et~al.(2022)]{brehmer2022weakly}
J.~Brehmer, P.~de~Haan, P.~Lippe, and T.~Cohen.
\newblock Weakly supervised causal representation learning.
\newblock In \emph{Advances in Neural Information Processing Systems}, volume~35, 2022.
\newblock arXiv:2203.16437.

\bibitem[Chen et~al.(2025)]{chen2025counterbench}
Y.~Chen, V.~K. Singh, J.~Ma, and R.~Tang.
\newblock {CounterBench}: Evaluating and improving counterfactual reasoning in large language models.
\newblock arXiv:2502.11008, 2025.

\bibitem[Gu et~al.(2024)]{gu2024webdreamer}
Y.~Gu, K.~Zhang, Y.~Ning, B.~Zheng, B.~Gou, T.~Xue, C.~Chang, S.~Srivastava, Y.~Xie, P.~Qi, H.~Sun, and Y.~Su.
\newblock Is your {LLM} secretly a world model of the internet? {M}odel-based planning for web agents.
\newblock arXiv:2411.06559, 2024.

\bibitem[Ha and Schmidhuber(2018)]{ha2018world}
D.~Ha and J.~Schmidhuber.
\newblock World models.
\newblock arXiv:1803.10122, 2018.

\bibitem[Hafner et~al.(2019)]{hafner2019planet}
D.~Hafner, T.~Lillicrap, I.~Fischer, R.~Villegas, D.~Ha, H.~Lee, and J.~Davidson.
\newblock Learning latent dynamics for planning from pixels.
\newblock In \emph{International Conference on Machine Learning}, 2019.
\newblock arXiv:1811.04551.

\bibitem[Hafner et~al.(2023)]{hafner2023dreamerv3}
D.~Hafner, J.~Pasukonis, J.~Ba, and T.~Lillicrap.
\newblock Mastering diverse domains through world models.
\newblock arXiv:2301.04104, 2023.

\bibitem[Hao et~al.(2023)]{hao2023rap}
S.~Hao, Y.~Gu, H.~Ma, J.~J. Hong, Z.~Wang, D.~Z. Wang, and Z.~Hu.
\newblock Reasoning with language model is planning with world model.
\newblock In \emph{Proceedings of the Conference on Empirical Methods in Natural Language Processing}, 2023.
\newblock arXiv:2305.14992.

\bibitem[Jin et~al.(2023)]{jin2023cladder}
Z.~Jin, Y.~Chen, F.~Leeb, L.~Gresele, O.~Kamal, Z.~Lyu, K.~Blin, F.~Gonzalez~Adauto, M.~Kleiman-Weiner, M.~Sachan, and B.~Sch{\"o}lkopf.
\newblock {CLadder}: Assessing causal reasoning in language models.
\newblock In \emph{Advances in Neural Information Processing Systems}, volume~36, 2023.

\bibitem[Ke et~al.(2021)]{ke2021systematic}
N.~R. Ke, A.~Didolkar, S.~Mittal, A.~Goyal, G.~Lajoie, S.~Bauer, D.~Rezende, Y.~Bengio, M.~Mozer, and C.~Pal.
\newblock Systematic evaluation of causal discovery in visual model based reinforcement learning.
\newblock arXiv:2107.00848, 2021.

\bibitem[Lachapelle et~al.(2022)]{lachapelle2022disentanglement}
S.~Lachapelle, P.~Rodr{\'i}guez~L{\'o}pez, Y.~Sharma, K.~Everett, R.~Le~Priol, A.~Lacoste, and S.~Lacoste-Julien.
\newblock Disentanglement via mechanism sparsity regularization: A new principle for nonlinear {ICA}.
\newblock In \emph{Conference on Causal Learning and Reasoning}, 2022.
\newblock arXiv:2107.10098.

\bibitem[Lippe et~al.(2022)]{lippe2022citris}
P.~Lippe, S.~Magliacane, S.~L{\"o}we, Y.~M. Asano, T.~Cohen, and E.~Gavves.
\newblock {CITRIS}: Causal identifiability from temporal intervened sequences.
\newblock In \emph{International Conference on Machine Learning}, 2022.
\newblock arXiv:2202.03169.

\bibitem[Madapathi(2026)]{madapathi2026zenodo}
M.~Madapathi.
\newblock Outcome-aware agents: From token prediction to action consequence modeling (first version).
\newblock Zenodo, 2026.
\newblock \url{https://doi.org/10.5281/zenodo.20379456}.

\bibitem[Pearl(2009)]{pearl2009causality}
J.~Pearl.
\newblock \emph{Causality: Models, Reasoning, and Inference}.
\newblock Cambridge University Press, 2nd edition, 2009.

\bibitem[Pitis et~al.(2020)]{pitis2020coda}
S.~Pitis, E.~Creager, and A.~Garg.
\newblock Counterfactual data augmentation using locally factored dynamics.
\newblock In \emph{Advances in Neural Information Processing Systems}, volume~33, 2020.
\newblock arXiv:2007.02863.

\bibitem[Ruan et~al.(2024)]{ruan2024toolemu}
Y.~Ruan, H.~Dong, A.~Wang, S.~Pitis, Y.~Zhou, J.~Ba, Y.~Dubois, C.~J. Maddison, and T.~Hashimoto.
\newblock Identifying the risks of {LM} agents with an {LM}-emulated sandbox.
\newblock In \emph{International Conference on Learning Representations}, 2024.
\newblock arXiv:2309.15817.

\bibitem[Sch{\"o}lkopf et~al.(2021)]{scholkopf2021causal}
B.~Sch{\"o}lkopf, F.~Locatello, S.~Bauer, N.~R. Ke, N.~Kalchbrenner, A.~Goyal, and Y.~Bengio.
\newblock Toward causal representation learning.
\newblock \emph{Proceedings of the IEEE}, 109(5):612--634, 2021.

\bibitem[Schrittwieser et~al.(2020)]{schrittwieser2020muzero}
J.~Schrittwieser, I.~Antonoglou, T.~Hubert, K.~Simonyan, L.~Sifre, S.~Schmitt, A.~Guez, E.~Lockhart, D.~Hassabis, T.~Graepel, T.~Lillicrap, and D.~Silver.
\newblock Mastering {A}tari, {G}o, chess and shogi by planning with a learned model.
\newblock \emph{Nature}, 588:604--609, 2020.

\bibitem[Wang et~al.(2022)]{wang2022cdl}
Z.~Wang, X.~Xiao, Z.~Xu, Y.~Zhu, and P.~Stone.
\newblock Causal dynamics learning for task-independent state abstraction.
\newblock In \emph{International Conference on Machine Learning}, 2022.
\newblock arXiv:2206.13452.

\bibitem[Zheng et~al.(2018)]{zheng2018notears}
X.~Zheng, B.~Aragam, P.~Ravikumar, and E.~P. Xing.
\newblock {DAG}s with {NO TEARS}: Continuous optimization for structure learning.
\newblock In \emph{Advances in Neural Information Processing Systems}, volume~31, 2018.

\end{thebibliography}
\end{document}